# Overview of Annotation Creation: Processes & Tools

Mark A. Finlayson and Tomaž Erjavec

**Abstract**
Creating linguistic annotations requires more than just a reliable annotation scheme. Annotation can be a complex endeavour potentially involving many people, stages, and tools. This chapter outlines the process of creating end-to-end linguistic annotations, identifying specific tasks that researchers often perform. Because tool support is so central to achieving high quality, reusable annotations with low cost, the focus is on identifying capabilities that are necessary or useful for annotation tools, as well as common problems these tools present that reduce their utility. Although examples of specific tools are provided in many cases, this chapter concentrates more on abstract capabilities and problems because new tools appear continuously, while old tools disappear into disuse or disrepair. The two core capabilities tools must have are support for the chosen annotation scheme and the ability to work on the language under study. Additional capabilities are organized into three categories: those that are widely provided; those that often useful but found in only a few tools; and those that have as yet little or no available tool support.

## 1 Annotation: More than just a scheme

Creating manually annotated linguistic corpora requires more than just a reliable annotation scheme. A reliable scheme, of course, is a central ingredient to successful annotation; but even the most carefully designed scheme will not answer a number of practical questions about how to actually create the annotations, progressing from raw linguistic data to annotated linguistic artifacts that can be used to answer interesting questions or do interesting things. Annotation, especially high-quality annotation of large language datasets, can be a complex process potentially involving many people, stages, and tools, and the scheme only specifies the conceptual content of the annotation. By way of example, the following questions are relevant to a text annotation project and are not answered by a scheme:

- How should linguistic artifacts be prepared? Will the originals be annotated directly, or will their textual content be extracted into separate files for annotation? In the latter case, what layout or formatting will be kept (lines, paragraphs page breaks, section headings, highlighted text)? What file format will be used? How will typographical errors be handled? Will typos be ignored, changed in the original, changed in extracted content, or encoded as an additional annotation? Who will be allowed to make corrections: the annotators themselves, adjudicators, or perhaps only the project manager?
- How will annotators be provided artifacts to annotate? How will the order of annotation be specified (if at all), and how will this order be enforced? How will the project manager ensure that each document is annotated the appropriate number of times (e.g., by two different people for double annotation).
- What inter-annotator agreement measures (IAAs) will be measured, and when? Will IAAs be measured continuously, on batches, or on other subsets of the corpus? How will their measurement at the right time be enforced? Will IAAs be used to track annotator training? If so, what level of IAA will be considered to indicate that training has succeeded?

These questions are only a small selection of those that arise during the practical process of conducting annotation. The first goal of this chapter is to give an overview of the process of annotation from start to finish, pointing out these sorts of questions and subtasks for each stage. We will start with a known conceptual framework for the annotation process, the *MATTER* framework (Pustejovsky & Stubbs, 2013) and expand upon it. Our expanded framework is not guaranteed to be complete, but it will give a reader a very strong flavor of the kind of issues that arise so that they can start to anticipate them in the design of their own annotation project.

The second goal is to explore the capabilities required by annotation tools. Tool support is central to effecting high quality, reusable annotations with low cost. The focus will be on identifying capabilities that are necessary or useful for annotation tools. Again, this list will not be exhaustive but it will be fairly representative, as the majority of it was generated by surveying a number of annotation experts about their opinions of available tools. Also listed are common problems that reduce tool utility (gathered during the same survey). Although specific examples of tools will be provided in many cases, the focus will be on more abstract capabilities and problems because new tools appear all the time while old tools disappear into disuse or disrepair.

Before beginning, it is well to first introduce a few terms. By *linguistic artifact*, or just *artifact,* we mean the object to which annotations are being applied. These could be newspaper articles, web pages, novels, poems, TV shows, radio broadcasts, images, movies, or something else that involves language being captured in a semi-permanent form. When we use the term *document* we will generally mean textual linguistic artifacts such as books, articles, transcripts, and the like.

By *annotation scheme*, or just *scheme*, we follow the terminology as given in the early chapters of this volume, where a scheme comprises a linguistic theory, a derived model of a phenomenon of interest, a specification that defines the actual physical format of the annotation, and the guidelines that explain to an annotator how to apply the specification to linguistic artifacts. (citation to Chapter III by Ide et al.)

By *computing platform*, or just *platform*, we mean any computational system on which an annotation tool can be run; classically this has meant personal computers, either desktops or laptops, but recently the range of potential computing platforms has expanded dramatically, to include on the one hand things like web browsers and mobile devices, and, on the other, internet-connected annotation servers and service oriented architectures. Choice of computing platform is driven by many things, including the identity of the annotators and their level of sophistication.

We will speak of the *annotation process* or just *process* within an annotation project. By *process*, we mean any procedure or activity, at any level of granularity, involved in the production of annotation. This potentially encompasses everything from generating the initial idea, applying the annotation to the artifacts, to archiving the annotated documents for distribution. Although traditionally not considered part of annotation *per se*, we might also include here writing academic papers about the results of the annotation, as these activities also sometimes require annotation-focused tool support.

We will also speak of *annotation tools*. By *tool* we mean any piece of computer software that runs on a computing platform that can be used to implement or carry out a process in the annotation project. Classically conceived annotation tools include software such as the Alembic workbench, Callisto, or brat (Day et al., 1997; Day, McHenry, Kozierok, & Riek, 2004; Stenetorp et al., 2012), but tools can also include software like Microsoft Word or Excel, Apache Tomcat (to run web servers), Subversion or Git (for document revision control), or mobile applications (apps). Tools usually have user interfaces (UIs), but they are not always graphical, fully functional, or even all that helpful.

There is a useful distinction between a tool and a *component* (also called an *NLP component*, or an *NLP algorithm*; in UIMA (Apache, 2014) called an *annotator*), which are pieces of software that are intended to be integrated as libraries into software and can often be strung together in annotation *pipelines* for applying automatic annotations to linguistic artifacts. Software like tokenizers, part of speech taggers, parsers (Manning et al., 2014), multiword expression detectors (Kulkarni & Finlayson, 2011) or coreference resolvers (Pradhan et al., 2011) are all components. Sometimes the distinction between a tool and a component is not especially clear cut, but it is a useful one nonetheless.

The main reason a chapter like this one is needed is that there is no one tool that does everything. There are multiple stages and tasks within every annotation project, typically requiring some degree of customization, and no tool does it all. That is why one needs multiple tools in annotation, and why a detailed consideration of the tool capabilities and problems is needed.

## 2  Overview of the Annotation Process

The first step in an annotation project is, naturally, defining the scheme, but many other tasks must be executed to go from an annotation scheme to an actual set of cleanly annotated files useful for other tasks.

### *2.1 MATTER & MAMA*

A good starting place for organizing our conception of the various stages of the process of annotation is the *MATTER* cycle, proposed by Pustejovsky & Stubbs (2013). This framework outlines six major stages to annotation, corresponding to each letter in the word, defined as follows:

**M = Model:** In this stage, the first of the process, the project leaders set up the conceptual framework for the project. Subtasks may include:
- Search background work to understand existing theories of the phenomena
- Create or adopt an abstract model of the phenomenon
- Define an annotation scheme based on the model

- Search libraries, the web, and online repositories for potential linguistic artifacts
- Create corpus artifacts if appropriate artifacts do not exist
- Measure overall characteristics of artifacts to ground estimates of representativeness and balance
- Collect the artifacts on which the annotation will be performed
- Track artifact licenses
- Measure various statistics of the collected corpus
- Choose an annotation specification language
- Build an annotation specification that distills the scheme and model
- Update annotation model and schemes on the basis of feedback from the *Annotate* stage
- Track differences between different versions of the models, schemes, and specifications

**A = Annotate:** This stage is the actual application of annotations to artifacts. Usually this stage involves multiple trained workers (annotators) who inspect the linguistic artifacts and decide which annotations are appropriate. Subtasks within this stage may include:

- Normalize artifacts, removing typos and other errors
- Create files in a standard file format
- Associate appropriate metadata with artifacts
- Write annotation guidelines
- Define necessary annotator skills and knowledge
- Recruit annotators
- Train annotators in the annotation workflow, including annotation tools to be used
- Train annotators in the scheme to reach an acceptable level of inter-annotator agreement (IAA)
- Plan the annotation order and assignments (respecting multilayer constraints)
- Distribute documents to the annotators
- Monitor annotator's progress
- Collect annotations from the annotators
- Ensure that annotation process metadata is captured (e.g., time to annotate, annotator identity, etc.)
- Track IAAs to ensure quality annotations
- Track annotation guideline versions
- Examine large sets of annotations for common errors or inconsistencies and apply corrections
- Update annotations in older versions of the specification to a new version
- Schedule annotator and adjudicator meetings
- Adjudicate multiple annotations into a gold standard
- Track worker hours and project budget
- Estimate artifact and corpus completion times

**T = Train & T = Test:** Pustejovsky & Stubbs were specifically interested in linguistic annotation for developing machine learning algorithms. In the second and third stage, therefore, they focused on training machine learning classifiers, and how to appropriately test them. This is a very important, yet specific application of linguistic annotation and is not always the goal of an annotation project. Researchers may, for example, be interested in just measuring a phenomenon of interest, validating some theory, or preparing data for others to use in their projects. Thus here we abstract away from the matter cycle a bit and replace 'TT' with **L = Leverage**. Namely, once you have the annotations, you should *leverage* them for your goal, be that training machine learning algorithms, manual inspection for testing linguistic theories, or something else.

**E = Evaluate:** No matter how you are planning on using your annotations, you should evaluate their utility for your purpose. In practice this usually involves one more or steps like:

- Explore and visualize the annotated data to get a qualitative sense of its scope, quality, and character
- Measure accuracy, precision, recall, or other statistics to numerically characterize the data
- Calculate confusion matrices, error classes, or other measures to categorically characterize the data

**R = Revise:** If the evaluation results are not satisfactory, one needs to revise some aspects of the annotation process. This is not a stage in and of itself, but acts more like an arrow pointing back to one of the previous steps.

Within this full cycle, Pustejovsky and Stubbs note that there is a subcycle, *MAMA*, which often happens at the beginning of an annotation project when you are still developing your model. This cycle involves iterating between modeling and pilot annotations, to increase the quality of the annotation scheme before investing the full amount of time and energy annotating the complete corpus. It is akin to developing a scientific hypothesis: you start by proposing a model, and then translating those into a specification and annotation guide. You train several annotators, have them annotate a small amount of data, and then inspect the data (either directly or with IAA measures). If the data fails inspection (e.g., you are missing a major category present in the data, IAAs are too low), you return to modify the model.

If the model itself is sound but the specification or guidelines fall short, there is an even smaller cycle that often takes place with the Annotation stage itself, whereby the guidelines are rewritten to be clearer. This cycle is illustrated in this volume (citation to Chapter V.f. by Artstein, Fig 1).

## 2.2 Additional Stages

While the MATTER framework is an excellent start, it still does not cover the full extent of an annotation project. We propose at least three additional stages: Idea, Procure, and Distribute.

**Idea:** Before creating the initial model, one must solidify one's question vis-a-vis existing linguistic knowledge and theory, plus have a rough idea of what language data might be used for the project. This may involve:
- Search the literature for concrete linguistic theories pertaining to the linguistic question of interest
- Verify that the phenomenon does not have an annotation scheme or annotated corpus that answers the question you are asking
- Explore existing corpora to determine if it might be profitable to annotate on top of those corpora
- Visualize existing corpora to determine if the information they contain is relevant to your question

**Procure:** After developing the model, but before beginning annotation, you must find the appropriate annotation tools for each task of the process. This stage may entail:
- Identify the various subtasks which follow from your annotation project design
- Identify tools that support these subtasks
- Identify tool capabilities that are critical to the project's success
- Obtain the tools that provide needed capabilities
- Modify existing tools to provide missing capabilities
- Create new tools that provide missing capabilities
- Verify that the tools work on the required computing platforms
- Verify that the tools can be assembled into a working annotation process
- Distribute patches and bugfixes for tools to annotators as they are working

**Distribute:** Once the data has been annotated and evaluated to the researcher's satisfaction, it is often the case that the research desires to distribute the data to the world at large. Although in-house, private corpora can be useful for certain limited pursuits, generally the best effect comes from a corpus when it is made available to the community. This stage may involve:
- Exporting the annotated artifacts from the annotation tool for distribution
- Cleaning the annotations of extraneous information
- Packaging annotated data and other material into downloadable or otherwise distributable archives
- Checking that artifact licenses are compatible with the planned distribution model
- Archiving data and other materials in a permanent archive (e.g., Institutional DSpace, LDC, etc.)
- Exporting data selections in publication-quality formats

The additional stages can be integrated into our abstract MATTER framework as shown in Figure 1.

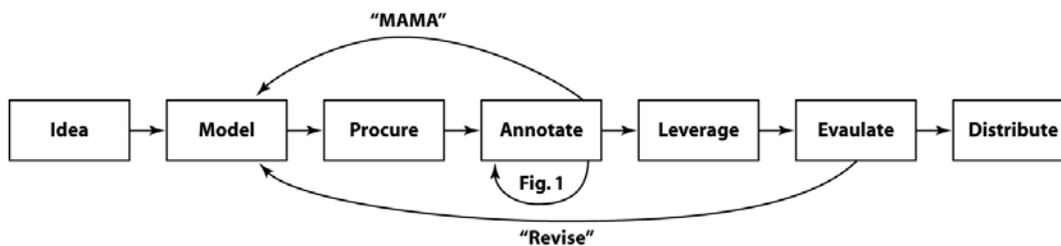

Figure 1: An abstracted and enhanced MATTER cycle. Figure 1 refers to that in Arstein (citation to Chapter V.f. by Artstein, Fig 1). Note that in course of looping you may naturally skip a number of steps. For example, you probably wouldn't re-procure tools within the MAMA subcycle as long as your technical requirements hadn't changed dramatically.

## 3 Basic Tool Considerations

The overview above listed seven stages of the annotation process, each with numerous subtasks. Each of the subtasks outside of the "Procure" stage is a candidate for annotation tool support. Usually you do not need as many tools as there are subtasks: often a single tool has the ability to perform many subtasks. Other subtasks you might accomplish without software support because it is easier or faster. On the other hand, you will most likely not be able to find a single tool that will handle all the required subtasks. This means you will need a number of tools to create your annotations.

The most important tool you choose is the one that provides the annotation user interface (AUI). This is the tool that annotators interact with to actually apply annotations to the linguistic artifacts. How intuitive, easy to use, and bug-free the AUI is has a direct and major impact on the speed and quality of the annotations. Moreover, the project often requires that the AUI have certain features, without which the project cannot proceed. Because of this centrality, usually the first major tool decision is to decide on the AUI. This will constrain a number of other decisions about how to carry out other subtasks of the annotation project. For example, the project might require crowdsourcing, in which case only AUIs that have this capability can be used. The project might involve annotating multimodal data such as video, audio, images, or combinations thereof: again, only certain AUIs can deal with this sort of data, and thus your choice is restricted to those tools.

Once the AUI is chosen this will also constrain what other tools you need and what other tools you can use. A particular AUI might only accept or export data in particular formats, meaning you will either have to live with those formats or transduce to and from them (citation to Wilcock chapter). A particular AUI might not implement certain capabilities (for example, version, document, user, or task control), and so you might have to adopt additional tools to provide those capabilities.

In this section and the next we discuss the requirements for the various tools involved in an annotation project, and how to choose between different tools, with a particular emphasis on capabilities that are usually found in the AUI.

### 3.1 Choosing the right tools

Choosing the right tools to accomplish an annotation project is not always a simple matter. Not only can each capability be accomplished with multiple tools, but each tool brings multiple capabilities to the table; so while tool X may be inferior to tool Y with regard to a particular capability C, tool Y might have some other capability D which tool X lacks, and which makes annotating with Y preferable to X on balance. Although small annotation projects may admit this strategy, it is not always as simple as ranking your desired capabilities, ranking tools according to their utility for those capabilities, and then proceeding down the lists in linear order.

As an example, consider the AUI, which the annotators use to mark up artifacts. Suppose the annotators are marking a TimeML time link scheme, which consists of marking "before" and "after" temporal relationships between events and times in a document. One might use brat (Stenetorp et al., 2012) for this, which provides a generic relationship annotation facility with an attractive UI that runs in a web browser. If the annotation project requires annotators to work remotely, with a variety of different computing platforms with only web browser access in common, brat may be the right tool. On the other hand, the TANGO tool (Verhagen, Knippen, Mani, & Pustejovsky, 2006) was specifically designed for TimeML and features optimized key bindings and UIs that present task-relevant information. TANGO also integrates automatic checks for potential annotation conflicts (where two individual annotations are not compatible). If the computing platform and project logistics allow it, it may be preferable to use TANGO for its specialization and additional features.

Another example would be access to external resources that help the annotator make efficient and accurate annotation decisions. A tool like MAE (Stubbs, 2011) can be used to apply an arbitrary tag to arbitrary spans of text. Such a tool can be used, for example, for Word Sense Disambiguation (Agirre & Edmonds, 2007), where an annotator associates a sense from a sense inventory—e.g., WordNet (Fellbaum, 1998)—with each open-class word. The user could have the WordNet dictionary open in a browser window, with MAE to the side, and copy and paste appropriate sense keys for the right words. In contrast, a specially designed tool like LX-SenseAnnotator (Neale, Silva, & Branco, 2015) might be preferred, as it integrates the WordNet database directly, and automatically identifies open-class words and provides only valid tags for the annotator to choose.

With these two examples in mind to remind us that a full ordering of tools does not exist, we can identify two main criteria for choosing one tool over another. The first, and rather obvious, capability is that the tools, especially the AUI, must support the chosen annotation schemes. The second most important capability is being able to work on the languages or character sets that the project aims to annotate. If you are annotating text in, say, Cyrillic, and the annotation tool cannot display that script, then you will not be able do your annotation. While most modern tools natively support Unicode, this is still not necessarily the case, especially for tools developed with only English in mind. However, even with character set support, there can be other language specific issues with tools. For example, some tools perform their own tokenization on the input texts, and if their tokenizer was developed for English, it will not work very well for other languages, say, Russian, and not at all for languages that do not use spaces to separate words, such as Japanese.

Beyond these primary capabilities, there are a host of secondary capabilities, although often no less critical to the success of the project. These are covered in detail below in §4. In the best case, then, you can find some combination of tools that can be brought together to provide all the needed capabilities in a single annotation process.

## 3.2 Creating the right tools

Sometimes the right tool to solve a particular task does not exist. Or, more often, a tool is mostly adequate, but is missing some key functionality or falls short in some other way. In these cases, researchers must either create a new tool from scratch or modify the existing tool to suit their needs.

The first option, and the one that usually occurs first to many researchers, is to build their own tool from scratch. Generally this route should be avoided. While the positive aspect is that you have complete control over the tool, there are many negatives. First, you will end up reinventing the wheel many times over, often less well than existing, vetted tools. Second, implementation decisions early on in the design process can hamstring the whole implementation and cause major headaches down the line. Third, it is a lot of work to create and maintain a tool: in the hands of your annotators (or even yourself), there will inevitably be bugs, and when these are found you must diagnose them, fix them, and distribute patches. Finally, if you release the tool for others to use, you will no doubt be subject to request for help from people using or trying to modify your tool.

In general it is better to find a tool that does most of what is needed and requires only minor modifications to add the missing functionalities. Examples of the sorts of capabilities that may need modification may be import or export to a particular file format, or visualizing or accessing some sort of external data. But beyond these rather broad capabilities, there might be something very simple that you need that the available tools don't provide. In these cases, it becomes important to consider two additional features of potential tools: extensibility and support.

**Extensibility:** Extensibility means the ability to add features to a tool that were not included in the original release. The conceptually simplest way to achieve modifications to a tool is to modify the source directly and recompile it, but of course the tool must be available in open source for this to be possible. When the tool is deposited on open repositories, such as GitHub, the extensions can furthermore be made available to the wider community.

A second type of extensibility is the use of a plugin architecture. For tools designed in this way you do not need to modify the original source code, but only parametrize to tool to make it aware of the additional code. A good example of this is GATE (Cunningham, Maynard, & Bontcheva, 2011).

In any event, modifying source code or creating plugins usually implies programming. This means that you must pay attention to the programming language the tool is written in and whether you have access to the expertise need to write the required code.

**Support:** Support refers to the resources available to help you understand how to modify the tool. Modifying the source code or creating a plugin necessarily requires you to understand in more detail the inner workings of the tool. Reverse engineering how a tool works by reading source code is time consuming, tedious, prone to error, and often leads to buggy code. Questions to ask yourself are: Does the tool have source code documentation? Are there manuals or guides that lead you through modifying the tool? Is there example code available that can serve as a template for the right approach? Is there a community of developers to which you can appeal for help if (or most likely, when) you get stuck? Better yet, can you contact the original developers to ask them questions? The less support there is, the harder it will be to carry out your modifications.

### *3.3 Common problems with tools*

Before moving on to the large number of capabilities that one might look for in annotation tools, it is worthwhile to consider a set of general problems that many tools present that can reduce their utility. A tool might in theory have a capability you need, but it might be so difficult to use or so buggy as to make it practically impossible to leverage that capability. Here we cover six of the most common complaints about annotation tools.

**Inadequate Importing, Exporting or Conversions:** A tool does not read or write to the formats you use, or doesn't understand the standards or schemes you want to use for your project. While there are many ways of transducing from one format to another (citation to Wilcock chapter), the format might be conceptually incompatible, which can be a serious detriment to a tool's utility.

**Lack of Documentation or Support:** Similar to lack of documentation for modifying tools, this problem refers to the tool lacking adequate documentation for those installing, administrating, and using the tool. The tool may provide a host of functions, but if you can't understand what the buttons do or the meaning of the menu items, then the tool will be much less useful. Having a well written user manual, an established user community to consult, or being able to ask questions of the original developers are major positives in a tool's favor.

**Difficult to Learn:** Even with proper documentation a tool can still be difficult to learn. Perhaps it uses unfamiliar or awkward user interface conventions. Or perhaps it organizes the workflow or functionality in a way which is unintuitive or doesn't match up well with the structure of your annotation project. This is a handicap.

**Poor User Interface:** A related problem to a difficult learning curve is a user interface (UI) that is just plain hard to use. Annotation is a repetitive task, and if an oft-repeated portion requires lots of work in the UI, then this can seriously impact the speed and quality of annotations. A similar problem can appear with Web-based platforms for annotation. If substantial data must pass between the browser and server for each key-stroke or mouse click, this can present latency problems, especially with slow internet connections.

**Difficult Installation:** Some tools require in-depth knowledge of the operating system or need a complex stack of technologies in order to install them successfully. In such cases a detailed installation guide and adequate support, either from the developers or from local system administrators, is a prerequisite for making the tool operational on a local machine.

**Unstable, Slow, or Buggy:** Finally, a tool that crashes a lot, takes a long time to do common tasks, or reports lots of errors is frustrating to use. As with documentation and support, production tools that have a large user base are much less likely to belong to this category than are experimental prototypes by a single author, which might run on their machine or for their project, but can cause problems in other environments.

## 4 Tool Features

Now that we have noted the primary capabilities of tools to support the annotation process, we move on to a broader array of features that support the various stages and subtasks of annotation projects. Not every annotation project will need all of these features, and their importance will vary depending on the project's goals. Therefore in this section we will divide features not by importance (which is variable), but by how easy it is to find tools that have the feature in question. This will hopefully assist you in prioritizing your effort in searching for the right tool for the job, versus spending time creating or new tool or modifying an existing one.

## *4.1 Common Features*

Features in this section are found in many different tools. They are also common across a range of annotation projects. The features here are listed in no particular order, and examples of a few tools that have the feature in question are provided. These lists of tools are by no means complete or exhaustive, and should not be taken as an endorsement of or recommendation to use that particular tool; they are merely well-known tools that have the feature.

**Importing/Exporting multiple file formats:** As all annotations are read from and written to files, the format of the files is clearly a consideration. The subfield or target audience may expect a particular file format (their analysis tools being written to accept that format), or the annotation project may build on other annotations or corpora which are provided in a specific format. In these cases one must be able to read annotations and data from the provided formats and write to the expected format. If the tool cannot do that, the project manager must transduce to and from the formats used by the tool (citation to Wilcock chapter). When designing your annotation workflow, consider carefully the various files you will be using, and whether your chosen tool can manipulate them without extra work.

Examples of tools that read and write multiple formats include: Praat, which is used for phonemic analysis, and reads numerous audio formats including wav, aiff, nist, and mp3, among others (Boersma, 2014); and GATE, which accepts and outputs a number of different types of text annotation formats including XML, UIMA CAS, CoNLL/IOB, and many others (Cunningham et al., 2011). Other tools that are notable for their choice of file formats include also ELAN (Auer et al., 2010), ExMARaLDA (Schmidt & Wörner, 2009), and WebAnno (to an extent) (Seid Muhie, Gurevych, Castilho, & Biemann, 2013).

**Standoff vs. inline annotation:** Aside from specific file formats, a more general consideration is whether the tool supports standoff or inline annotation. In standoff annotation, the original artifact is not modified, rather, annotations are stored in separate documents and associated with the artifact by means of pointers into the artifact. For example, annotations of a text file might be associated with a particular span of characters indicated by start and end character counts. Annotations of an audio file might be associated with a time span delimited by start and end times. Inline annotation, in contrast, inserts annotations directly into the artifact being annotated. Later tools that read the artifact must then be sensitive to these annotations so that they may be used or ignored as appropriate. Examples of this include the common format of "token/pos-tag" (e.g., "The/DT dog/NN ran/VB ./.") or the CoNLL format (Buchholz, Marsi, Krymolowski, & Dubey, 2015).

In any case, standoff annotation is usually considered a best practice, and so using a tool which provides this capability is usually preferred. There would be cases, though, in which the ability to produce inline annotations could be useful, such as when you are using later text processing tools that require inline annotations as input.

Tools that do standoff annotation include most modern tools like MAE, brat, and WebAnno (Seid Muhie et al., 2013; Stenetorp et al., 2012; Stubbs, 2011). Older tools like Alembic and Callisto (Day et al., 1997, 2004) usually produce inline annotations. GATE can read and write annotations in a selection of both standoff and inline formats. Like file formats, it is possible to use external transducers to translate between standoff and inline, however, this is in general complicated as inserting standoff annotations can break well-formedness of the document, by introducing so-called crossing hierarchies, where inserted standoff annotations do not nest properly with the original annotations; this is not allowed in, e.g., XML.

**Multi-layer annotation:** In the past many annotation projects involved adding only a single type of annotation to artifacts. As NLP progressed and more annotated data and annotation schemes became available, more and more projects added multiple types of annotations to artifacts. The first case is called single-layer annotation, and the second case is called multi-layer annotation. This capability also impacts the choice of standoff versus inline, as multilayer annotations are usually best expressed as standoff annotations, also because of the problem of crossing hierarchies mentioned above.

Another consideration here is whether the tool allows multiple layers to use the same annotation scheme. For example, does the tool allow two different, non-interacting layers of part of speech tags? There are times when this capability can be useful, such as when inspecting common semantics between two different schemes, doing annotation adjudication, or performing comparisons of different analyzers that produce the same type of tag.

Examples of tools that provide multi-layer annotation capabilities include MAE (Stubbs, 2011), GATE (Cunningham et al., 2011), and the Story Workbench (Finlayson, 2011).

**Multimodal annotation:** Multimodal annotation refers to the ability to annotate artifacts that contain multiple *modalities* of data, such as text and speech, or audio and video. Sometimes multimodal is used to refer to artifacts that just contain a modality other than text (which is especially easy to visualize). To annotate multimodal artifacts, one needs much more complex visualizations. A nice example is in Praat, where it is often necessary to visualize the spectrogram, tone level, and transcription of an audio file, all time aligned. In the case of true multimodal artifacts, one must have visualizers for each modality plus often some way of visualizing the alignment between modes.

Common tools that were purposely built to support multimodal annotation include Praat for phonemic annotation (Boersma, 2014), ANVIL for video annotation (Kipp, 2014), and CLAN for transcription (MacWhinney, 2015).

**Annotation-Customized UI:** As has been noted, a user-friendly UI for one's tools is an important feature. More specifically with regard to the AUI (the tool that is actually used by the annotators to do the annotation), a feature of great value is a UI that has been customized for annotating the chosen scheme. There is a significant difference between a UI that can, in theory, allow a particular scheme to be annotated and a UI that is specifically optimized to allow efficient annotation of the scheme with a minimum of error. Optimization can be as simple as bringing key menu items to the fore, highlighting particular buttons of use at a particular stage, or providing keyboard shortcuts for the most often used operations.

Two examples of tools that are optimized with respect their particular annotation schemes are brat and TANGO. Brat (Stenetorp et al., 2012) is specifically optimized to annotate and visualize sparse, local relations in text, such as events or dependency structures. It provides a simple, intuitive mouse-click-driven interface that allows an annotator to quickly create and label relations between text spans according to a specified relation schema. It also provides a good example of how an interface optimized for one task can quickly become a burden even for closely related tasks. For example, while brat excels at sparse local relations, it falls short for annotation schemes that beget extremely dense relations, or relations that span text beyond one or more lines. In these cases the brat user interface quickly becomes cluttered and confusing.

TANGO (Verhagen et al., 2006) is another example of a tool optimized for a particular annotation task, in this case, annotating TimeML relations. Additional examples are Palinka, which can be used for co-reference annotation (Orasan, 2001), or Jubilee, which was specifically designed to efficiently annotate the PropBank standard for Semantic Roles (Choi, Bonial, & Palmer, 2009).

**Agreement calculations:** Calculating agreement between sets of annotations applied to the same artifact by different annotators is a fundamental operation involved in vetting annotated corpora and ensuring their quality. Most peer-reviewed reports on corpus contents are required to include measures of inter-annotator agreement (IAA). While numerous external tools (such as MATLAB, R, or generic programming language environments) can be used to do IAA calculations, it is quite useful when the AUI or other annotation-related tools provide this service. Examples of tools that provide IAA measurement include WebAnno (Seid Muhie et al., 2013), GATE (Cunningham et al., 2011), and Djangology (Apostolova, Neilan, An, Tomuro, & Lytinen, 2010).

**Adjudication interface:** Related to IAA calculation is a tool that allows an adjudicator to quickly and easily merge annotations from different annotators to produce a gold standard. This can be a tricky task, with quite a bit of difficulty in visualizing the differences between two annotations. As with the AUI itself, the efficiency of the adjudication interface has a dramatic impact on the productivity of the adjudicator. For example, the simplest approach to adjudication is just to open two separate AUI instances which show the two different versions, informally designating one file as the master copy and another as the secondary. But this approach is awkward, requiring the adjudicator to switch their attention from one window to the other at quite a distance apart on the screen, identifying subtle differences between annotations without any visual highlighting or other aids. Furthermore the adjudicator must manually copy over information from the secondary to the master, which provides many opportunities to introduce errors.

A good example of a tool specifically tailored to adjudication is MAI (multiple document adjudication interface) (Stubbs, 2011). MAI uses different user interface colors to indicate different types of inter-annotator disagreements, and allows the adjudicator to correct tags individually or add new tags to the gold standard; the tool demonstrates how a specifically tailored adjudication interface can significantly streamline the process of producing a gold standard. WebAnno also provides this capability (Seid Muhie et al., 2013).

**Capturing metadata**
An often overlooked task in annotation projects is capturing metadata about the annotation process itself. Depending on how fine-grained the metadata is that is required by the annotation project, this may require some sophisticated integration with the AUI. At a bare minimum, one usually wants to know which annotator annotated which document, and which documents were already annotated. But an annotation project manager may be interested in more detail, for example, such as how long an annotator worked on a particular document or the provenance of an individual annotation: how was it originally generated, in what order was it modified, by whom, and how? This information can be used to analyze the annotation process for later improvements, or measure annotator productivity and efficiency.

A tool that integrates a sophisticated metadata capture system is the Story Workbench (Finlayson, 2011), which captures both annotation provenance and annotation timing data at the level of the individual annotation.

**Corpus analytics & pattern analysis:** A common task when starting a linguistic annotation project is to characterize the corpus to be annotated. For text, it is not uncommon, for example, to count various document or token types, or characterize the vocabulary. Key Word in Context (KWIC) analyses can also be useful, especially when inspecting the semantics of individual words. EMU, used for annotating speech, is an example of a tool that provides this sort of functionality (Bombien, Cassidy, Harrington, John, & Palethorpe, 2006), relying on its close integration with the R programming environment to allow the calculation of sophisticated statistics of corpus contents. Beyond this tool very few AUIs integrate this functionality directly. There are, however, stand-alone tools that provide it which may be brought to bear on the problem, for example, the Sketch Engine (Kilgarriff, 2014). Of course, use of an external tool like this implies the problem of importing corpus data to the tool for analysis.

**Creating arbitrary flat tag schemes:** An extremely common, even prototypical, linguistic annotation scheme structure involves defining a set of tags that are to be applied to spans of text (citation to Chapter III by Ide et al,). This type of annotation project is so common that tools that provide the ability to define an arbitrary tag scheme and apply it to data can be immediately useful to a wide range and variety of annotation projects. Considerations here involve what constraints the tool places on how the scheme is defined: are there restrictions on the types of tags? What UI elements are used to choose tags? Can the scheme designer restrict what spans of text may be annotated on the basis of other information (tokens, sentences, paragraphs, can't cross sentence boundaries)?

Tools that provide the ability to define and then annotate with an arbitrary scheme are fairly common, as this feature has been found in AUIs since the early days of Callisto and Alembic (Day et al., 1997, 2004). Modern standout examples include Ellogon (Petasis & Karkaletsis, 2002), MAE (Stubbs, 2011), WebAnno (Seid Muhie et al., 2013), and GATE (Cunningham et al., 2011).

**Web-based annotation:** In today's increasingly web-interconnected world, the ability to perform and collect annotations via a browser-based interface on a centralized platform is becoming a commonly desired feature. Embedding an AUI in a browser-based application or webpage has several advantages: it does not require the annotator to install anything (except a browser, which most already have), it allows recruitment of annotators far and wide, and it allows annotators to work remotely. Example tools that have a centralized server with web interfaces that are quite functional include brat (Stenetorp et al., 2012), WebAnno (Seid Muhie et al., 2013), and EXMARaLADA (Schmidt & Wörner, 2009).

**Access to external resources:** As noted previously, the ability to access external resources such as electronic dictionaries, theasauri, or knowledge bases (ontologies) can be a key capability for many annotation projects. The annotators might need to reference the resource to make annotation decisions (for example, searching for a particular word or concept). The more closely such functionality is integrated with an AUI, the easier it usually is for the annotator to take advantage of the resource. Other projects require the direct application of items from the resource to the artifacts. A good example is Word Sense Disambiguation, which requires the annotator to pick a sense present in the electronic dictionary (such as WordNet or the LDOCE) and associate it with the word.

Examples of tools that bring in external resources for reference include Jubilee for VerbBank (Choi et al., 2009), LX-SenseAnnotator for WSD (Neale et al., 2015), or the Story Workbench (Finlayson, 2011), which provides access to WordNet, the PropBank frame library, and VerbNet.

## 4.2 Uncommon Features

In contrast to the features and capabilities in the previous section, there are a number of features that annotation projects often need, but are not commonly found in AUIs or other annotation support tools.

**Creating arbitrary annotation schemes:** In the list of common features above we included creating flat tag schemes. As noted in a previous chapter, annotation schemes come in different types, including single labels, sets of "flat" attribute-value pairs, full-fledged recursive feature structures, relations between segments, or some combinations of these . (citation to Chapter III by Ide et al.). Although there is plenty of support for defining and annotating single label tagsets, and brat (Stenetorp et al., 2012) allows definition of arbitrary relation schemes, there is little or no support for defining the other more complicated types of annotation schemes such as recursive feature structures or combination schemes. Thus if one's annotation scheme involves any of these more complicated structures, one is almost forced to modify an existing tool or create a new tools. This is a major limitation of annotation tools, especially as the field moves toward more complicated linguistic phenomena.

One example of a tool which does include this functionality is SALTO, which allows dynamic definition or extension of an annotation scheme by adding new frames, frame elements, and flags (Burchardt, Erk, Frank, Kowalski, & Pado, 2006).

**Sophisticated Visualization:** A useful feature, but one not often found in AUIs focused on text alone, is that of sophisticated visualization of annotations. Some annotations schemes can be quite complicated, involving large tag sets, numerous types of linguistic objects, and multiple features arranged into complicated hierarchies. Visualization can thus be of great service in understanding the current state of the annotation of the document, perceiving errors, and determining what needs to be done. Moreover, annotation, as has been noted several times already, can often be tedious and somewhat mind-numbing for the annotator. This has the effect of making it easy for annotators to miss key pieces of information; thus, an AUI that visualizes annotations in an intuitive, clear, and expressive manner helps to increase the efficiency of annotators and the quality of their annotations.

Multimodal tools (such as Praat, ANVIL, and CLAN) tend, by the very nature of their targeted linguistic artifacts, to have sophisticated visualization facilities. Tools for the text annotation, on the other hand, often lack comparable visualization capabilities that truly take advantage of the full power of a modern graphical user interface. Notable exceptions include brat (Stenetorp et al., 2012), which excels at visualizing sparse local relations, and ANNIS which does not provide annotation capabilities *per se* but rather specializes in search an visualization of annotations (Zeldes, Ritz, Lüdeling, & Chiarcos, 2009).

**Checking file correctness against specs:** It is often of great utility to be able to verify that an annotated artifact conforms to some specification of the format of the annotations. For example, that the tagset used is the one claimed, with no extra or misformatted tags. This is akin to verifying that an XML document is valid, so, not only syntactically well-formed (e.g., all opening tags have a corresponding closing tag, tags properly nested as a tree), but that it also follows the required grammar of the tags as specified by an XML schema. One example is CLAN, which provides the ability to check if a particular annotation file conforms to the CHAT annotation file format. And, moreover, any tool that can import a particular format performs an implicit well-formedness check, in that if the import of a particular file succeeds you can be sure that the file conforms at least to that particular tool's implementation of the format specification. But the more explicit and general form of this feature is desirable: being able to affirm (without the tool crashing or producing some other error behavior) that a file is formatted correctly relative to some formal specification, and contains neither formatting errors nor extraneous unformatted material.

**Workflow support (user, role, file, and task management):** A fairly important but often overlooked set of capabilities, especially from the point of the view of an annotation manager, is the ability to manage the overall workflow of an annotation project. By *workflow* in this case we mean the process of planning the unfolding of an annotation project in terms of individual tasks, annotators, and files. What files will be annotated at what time, and by whom? Are there constraints that must be satisfied (i.e., one annotator must annotate first, or one file must be annotated before another)? If task assignment is unconstrained, and annotators are allowed to pick and choose what files they do when, how will you assure that they only annotate a file once, or do not annotate files they are not supposed to, or do not miss a file? Moreover, how exactly will files be distributed to annotators and the annotators notified of their assignments? By email? Shared file system? Other network file distribution facility?

Examples of tools that support such features, including fine-grained control over user access rights and file and task assignments, include the LDC tools suite, SALTO and WebAnno. The LDC tools (which are, to our knowledge, not generally available), allow flexible assignment of annotation tasks to geographically spread-out annotators; the development of that suite was driven by the large-scale and time-sensitive nature of many of LDC annotation projects. SALTO gives the ability to assign files for annotation to one or more annotators within a special administrative mode (Burchardt et al., 2006). The WebAnno editor allows defining a pool of annotators and files for a project, distributing the files among the annotators and monitoring progress (Seid Muhie et al., 2013). A few other tools also provide related capabilities (e.g., Anafora: Chen & Styler, 2013), but generally workflow management is under-attended to.

**Customizable annotation pipeline:** Annotation today is becoming more and more of a sequenced affair. That is, instead of starting with a plain, unannotated linguistic object, annotation projects will often rely on applying a number of automatic layers of annotation before beginning their own annotation. For text, good common examples of types of processing applied to text before more high-level annotation takes place include tokenization, sentence segmentation, part-of-speech tagging, lemmatization, and syntactic parsing. In these cases, it very helpful if the tools used to create the files for annotation allow the assembly or arbitrary automatic annotation pipelines. If these processing capabilities, however, are not integrated with an AUI, such as with WebLicht (Hinrichs, Hinrichs, & Zastrow, 2010), then an extra step of transferring files from the pre-processing pipeline to the AUI must be undertaken.

A bare bones example of a fully customizable processing pipeline is something like UIMA (Apache, 2014), which allows assembling arbitrary sequences of automatic annotators using a number of different programming languages. This situation is not necessarily ideal, however, as the learning curve for UIMA is a bit difficult and requires some sophisticated programming skills. Good examples of tools that provide a reasonable UI to create pipelines but still allow sophisticated pre-processing of text files include GATE (Cunningham et al., 2011) and WebLicht (Hinrichs et al., 2010). On the other hand, more and more annotation platforms do integrate the ability to automatically pre-annotate files, at least for low-level annotations. Most of them have the annotation program built-in, which means it works only for particular types of annotation and particular languages. Some others, in particular WebAnno incorporate a generic machine learning program, that allows the administrators to define the type of annotation to be performed, import training data, and train the learner on this data. New data can then be automatically annotated with the trained model.

**Interleaving manual and automatic annotation:** Related to the issue of assembling annotation pipelines and online learning is interleaving manual and automatic annotation. Sometimes is useful to have a tighter feedback loop between manual and automatic stages of the annotation process: do some pre-processing annotation, have annotators correct or add to those annotations, and then do more automatic annotation. When returning to the automatic stage, the automatic analyzers take advantage of the cleaner and corrected manual annotations so as to do a better job themselves.

An example of a tool that interleaves these two modes in a smooth way is the Story Workbench (Finlayson, 2011). When an annotator modifies a file, usually by correcting or adding an annotation, the Story Workbench calculates the changed portion of the text and re-runs the automatic analyzers that are set up to run on that file. The difficulty with that implementation, however, is that, unlike UIMA or GATE, the processing sequence is not especially flexible.

**Online learning:** At the far end of the spectrum of integration of automatic and manual annotation is online learning. This was a feature found in the very earliest AUIs such as the Alembic workbench (Day et al., 1997). In this approach the system is constantly observing the annotator's actions, and retraining a model that drives an automatic annotator. After each retraining the system retags everything that has not yet been touched by the annotator.

Of modern tools, the only one we know of that implements this quite useful feature is Annotate (LREC 2000), which provides incremental annotation of context-free grammar analyses. Another tool that implements this functionality is CorA (Marcel Bollmann, Florian Petran, Stefanie Dipper, 2014), which can use the manually annotated data, possibly in combination with pre-existing annotated corpora, to train its normalizer and tagger. The tool can also be extended with PHP classes to add further online learning modules, e.g., lemmatization or sentence boundary detection.

**Crowdsourcing:** Of increasing interest lately is the opportunity to conduct annotation through crowdsourcing; using online work distribution platforms like Amazon's Mechanical Turk or Crowdflower. The appeal in these cases is easing the recruitment of annotators and quickly scaling up annotation projects at low cost. The difficulties include integrating the chosen crowdsourcing platform into the project's workflow (e.g., transferring data in and out, tracking progress), and providing annotators with the appropriate training and AUI to perform the annotation. While this capability is in high demand right now, there are few integrated solutions available. Two examples are the GATE (Cunningham et al., 2011) and WebAnno crowdsourcing plugins, (Seid Muhie et al., 2013, sec. 3.1.6), both of which interface with the Crowdflower platform.

**Querying:** Like any complicated set of data, the ability to search for specific pieces of information parameterized along dimensions of relevance to the data is a general ability of great use to many other tasks. This is more than just being able to search for specific spans of text of the presence of individual tags. One might want to formulate structured queries, such as "find all annotations which have a tag at this point in their structure", or "find all annotations across the whole corpus which have feature X and occur just before another annotation with feature Y." Although basic search abilities are quite common, these more complex search abilities keyed to the annotation schemes themselves are somewhat rare. Emu (Bombien et al., 2006) integrates a good facility for searching in this manner, with the ability to search provided by the Annotation Graph API (Maeda, Bird, Ma, & Lee, 2002).

### *4.3 Missing Features*
Finally, there are a number of features that very seldom have good tool support with AUIs or other tools designed for annotation. Annotation project managers must either do without a fully functional support for these features or "roll their own" solution, making use of *ad hoc* collections of tools and procedures.

**Ability to correct the original artifact:** With linguistic artifacts it is not uncommon to uncover errors. These may be of a typographical nature (such as a misspelled word or incorrect punctuation), or more like transduction errors, such as in the case of transcription which should reflect an underlying audio file. In these cases, it is extremely useful to be able to correct the original artifact. It is best, naturally, to discover and correct these errors before annotation begins. In practice, however, annotators will usually find overlooked errors. There are two issues of concern. The first is whether annotators should be allowed to make corrections themselves, and if so, how the project manager will keep track of the correction and how they will be propagated to other annotators working on the same file. The second issue, in the case of stand-off annotation, is that one must ensure that this modification does not make the indices of existing annotations invalid. Generally, support for correcting the original artifact is lacking. However, there are isolated tools, such as CorA, that do provide this support, as well as for the related task of correcting tokenization errors, tokens often being taken as the basic units over which annotation is indexed. CorA was specifically designed to annotate historical texts, where transcription errors are quite common, and allows correcting, deleting and inserting tokens in the primary data, and supports token level annotations for normalized and modernized word form, their lemma, part-of-speech and morphological feautres (Marcel Bollmann, Florian Petran, Stefanie Dipper, 2014).

**Annotation error detection and correction:** Error detection is related to querying. In the course of an annotation project being able to automatically detect and correct errors is useful, especially in the early and late stages. In the "MAMA" stage of the annotation project one is repeatedly examining small batches of annotations for errors, finding patterns, and then returning to either rewrite the annotation guidelines, retrain the annotators, or rework the annotation model. At the end of the project, when the data is fully annotated, one goes through the same procedure, but this time is usually looking for specific inconsistencies identified in the course of annotation, and quickly applying a large number of corrections. Some interesting work in this area has been done under the auspices of the DEECA project (Dickinson & Lee, 2008), but much of this work has not yet found its way into existing annotation tools.

**Annotation scheme editor:** Another oft-needed feature is the ability to edit annotation schemes and specifications via a dedicated user interface, rather than editing them directly in the file. While a number of tools mentioned above allow project managers to use their own customized annotation scheme, the tools have minimal to no support for actually creating the custom scheme. Usually it is assumed that the scheme will be created in a text editor, or, at best, an XML editor. Much like how integrated development environments with specialized editors ease computer programming, so too would specialized editors for annotation schemes. Such editors would support defining new schemes, extending existing schemes, and checking schemes for correctness and compatibility with known schemas. A tool that does offer this support is WebAnno (Seid Muhie et al., 2013), which allows the creation of new annotation layers, which can be either per-token annotations, with or without a predefined set of values, span annotations, and arc annotations for, e.g., co-reference or dependency annotations.

**UI builder:** Related to the ability to create annotation schemes, another extremely useful feature would be the ability to customize a user interface for annotating a particular scheme. Here we think of classic GUI builders available for the window toolkits for various programming languages. The ability to optimize a user interface by defining window component locations and sizes, menu structures, and keyboard shortcuts would go a long way toward allowing project managers to adapt existing AUIs to new annotation projects.

**Managing specifications, guidelines, and corpus versions:** This capability refers to the ability to manage and work with, simultaneously, many versions of the same annotation object. Over the course of an annotation project, things like the annotation scheme specification and annotation guide go through several versions. As a new version of the specification is applied the portion of the corpus with the old specification, it is likely that both the new and the old versions will co-exist simultaneously. AUIs would do well to support this, showing clearly which version is in use at any given time, allowing annotators to see the differences between two versions of a scheme or guideline.

**Managing and measuring annotator training:** Every annotation project requires training annotators. Sometimes this happens in a fully face-to-face manner; in the case of crowdsourcing projects all training might be done remotely; sometimes it is a mix of the two. Furthermore, sometimes training is extensive (weeks, with continuous testing and re-training), sometimes it is a matter of a few sentences of instruction. As training becomes more complicated, remote, and lengthy the more useful a facility to manage annotator training becomes. Such a capability would at a minimum allow assignment of training texts and measurement of annotator agreement against a gold standard (sometimes available in tools); in the ideal case such a system would be able to provide targeted feedback to an annotator about common mistakes, pointing them to key examples or portions of the annotation guide or possibly weeding out untrustworthy annotators

**Support for packaging into archives, distributing to repositories, and managing licenses:** Another under provided feature is some uniform ability to package annotated corpora, publish them to a permanent repository, and keep track of the licensing schemes for the data. Most annotation projects today make very simple use of external file formats like .zip or .tgz to package and distribute data. At best, projects will place their corpora in permanent archives like LDC or, in Europe, the CLARIN repositories. To understand how this process might be improved upon, it is instructive to look at the case of Maven, which is a relatively recent development in the world of automatic build tools for Java. Maven provides a centralized repository for Java code libraries, with standardized names and packaging conventions for things like source code, code documentation, licenses, test code, binaries, and so forth. The Maven build tools take a standardized package description that shows the tool where all these different pieces may be found, and the tool can communicate directly with the central repository and publish artifacts to it, whereupon they are immediately available to all other users of Maven. An analogous system for annotated corpora could potentially be quite useful.

**Exporting to publication-quality formats:** Finally, because much annotation work finds its final transmission and description in published works such as journal articles, conference papers, and books, a facility for transforming annotated data into publication quality figures would be quite useful. This is especially the case when the script of the linguistic object is complex and difficult to produce, or when the annotations are complex or hierarchical.

## 5 Conclusion

In this chapter we have reviewed the general process of annotation, identifying the general stages and subtasks to be found in each. We outlined several common problems, and then listed numerous features that are useful for carrying out an annotation project.

For each case we gave examples of individual tools that provide the features discussed. It is important to remember that these lists of example tools are neither complete nor exhaustive; and the mention of a particular tool should not be construed as a recommendation of that tool over other tools that may share the feature. As discussed, choosing a tool or set of tools (especially the AUI) is governed by project-specific considerations that dramatically change the desirability of various tools.

Further, we have reviewed many features. But it is important to remember that most annotation projects will not need all these features; indeed, most projects will only have a critical need for a handful. Don't be tyrannized by choice: identify the absolutely most important features and let those guide you. In most cases, this will be enough to determine your tool choice. The many additional considerations mentioned here can be appealed to in those rare cases when there are multiple tools that can actually do the job.

Finally, having read through this article, a reader might find himself discouraged from pursing annotation altogether: perhaps it is too complicated and difficult to do correctly. It is not our intent to give this impression at all. Indeed, small annotation projects can often be pulled together with a minimum of time and effort. With some thought, and early consideration of the issues discussed in this handbook, the researcher new to annotation can avoid the most common pitfalls and produce quality data on the first try.

Overview of Annotation Creation: Processes & Tools        17